\pdfoutput=1

\documentclass[11pt]{article}

\usepackage{acl} 

\usepackage{times}
\usepackage{latexsym}
\usepackage{framed}
\usepackage{amsmath} 
\usepackage{amssymb}
\usepackage{graphicx}
\usepackage{url}
\usepackage{tabularx}
\usepackage{booktabs}
\usepackage{multirow}
\usepackage{soul}
\usepackage{subcaption}

\usepackage[T1]{fontenc}

\usepackage[utf8]{inputenc}
\usepackage{microtype}

\newcommand*{\affaddr}[1]{#1}

\title{Discord Questions:\\ A Computational Approach To Diversity Analysis in News Coverage}

\author{
  \quad \textbf{Philippe Laban $^{\diamondsuit}$}
  \quad \textbf{Chien-Sheng Wu $^{\diamondsuit}$}
  \quad \textbf{Lidiya Murakhovs'ka $^{\diamondsuit}$} \\
  \quad \textbf{Xiang `Anthony' Chen $^{\clubsuit}$}
  \quad \textbf{Caiming Xiong $^{\diamondsuit}$} \\
  \affaddr{$\diamondsuit$ Salesforce AI Research, $\clubsuit$ UCLA} \\
  \{plaban, wu.jason, l.murakhovska, cxiong\}@salesforce.com \\
}

\begin{document}
\maketitle
\begin{abstract}
There are many potential benefits to news readers accessing diverse sources. Modern news aggregators do the hard work of organizing the news, offering readers a plethora of source options, but choosing which source to read remains challenging.
We propose a new framework to assist readers in identifying source differences and gaining an understanding of news coverage diversity.
The framework is based on the generation of \textit{Discord Questions}: questions with a diverse answer pool, explicitly illustrating source differences.
To assemble a prototype of the framework, we focus on two components: (1) \textbf{discord question generation}, the task of generating questions answered differently by sources, for which we propose an automatic scoring method, and create a model that improves performance from current question generation (QG) methods by 5\%, (2) \textbf{answer consolidation}, the task of grouping answers to a question that are semantically similar, for which we collect data and repurpose a method that achieves 81\% balanced accuracy on our realistic test set.
We illustrate the framework's feasibility through a prototype interface. Even though model performance at discord QG still lags human performance by more than 15\%, generated questions are judged to be more interesting than factoid questions and can reveal differences in the level of detail, sentiment, and reasoning of sources in news coverage. \footnote{Code is available at \url{https://github.com/Salesforce/discord_questions}}.
\end{abstract}

\section{Introduction}

\begin{figure}
    \centering
    \includegraphics[width=0.43\textwidth]{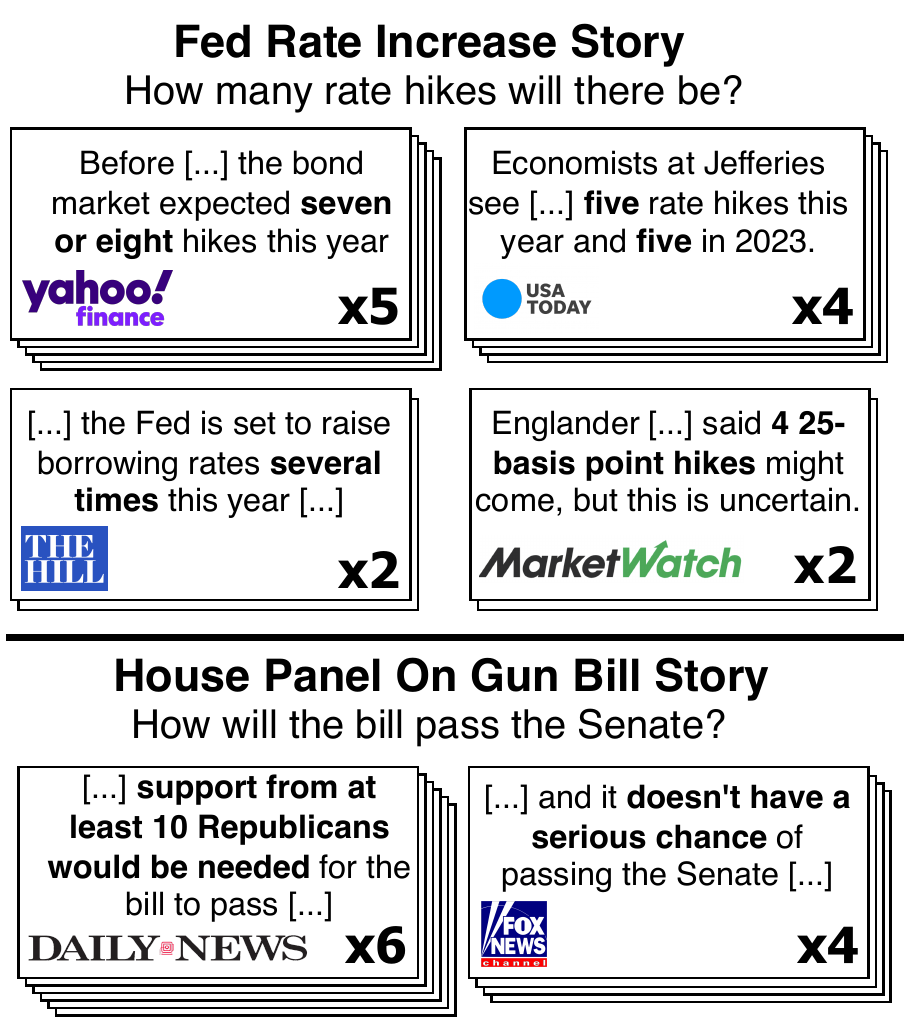}
    \caption{\textbf{Discord Questions surface news coverage diversity.} By finding questions that sources answer differently, concrete examples of coverage diversity for a particular story can be surfaced.}
    \label{fig:high_level_examples}
\end{figure}

News coverage often contains bias linked to the source of the content, and as many readers rely on few sources to get informed \cite{newman2021reuters}, readers risk exposure to such bias on critical societal issues such as elections and international affairs \cite{bernhardt2008political}.
Modern news aggregators such as Google News propose an engineering solution to the problem, offering news readers diverse source alternatives for any given topic.
In practice, however, users of news aggregators interested in diverse coverage must invest more time and effort, reading through several sources and sifting through overlapping content to build an understanding of a story's coverage diversity.

Prior work has explored methods to present coverage diversity information. For example, AllSides offers meta-data about the sources, such as political alignment \cite{allsides2021allsides}. But source-based information can be overly generic. Other projects have proposed to use article clustering and topic-modeling-based approaches to provide the user with story-specific insights about source diversity. Yet clustering interpretation can be complex for untrained users \cite{Spinde2020EnablingNC, saisubramanian2020balancing}. 

In this work, we propose a new framework to discover and present news diversity in multi-source settings: the \textbf{Discord Questions} framework. Discord questions are meant to be: (1) answered by most sources that cover the story, (2) answered in semantically diverse and sometimes contradicting ways by the sources. The use of questions to accompany readers is motivated by prior work showing automatically generated questions can improve reader comprehension \cite{therrien2006effect}, and foster an environment for active reading and comprehension \cite{singer1978active}.

The discord questions and the consolidated groups of answers are intended to be an interpretable slice through the sources' coverage, indicating how sources align for a specific issue in the story. Figure~\ref{fig:high_level_examples} presents two illustrative discord questions that were generated by our framework existing Google News stories.
In the first example, the sources and experts they introduce make forecasts that are subjective and uncertain: in a story about the Federal Reserve's rate increase, news sources predict that anywhere between 4 and 8 hikes might happen in 2022.
In the second example, in a story about the US House passing a bill about Gun Regulations, some sources chose to be more optimistic, focusing on how many Republicans were required for the bill to pass, while others employed a more pessimistic tone, writing that the bill did not have a serious chance to pass.

\begin{figure}
    \centering
    \includegraphics[width=0.47\textwidth]{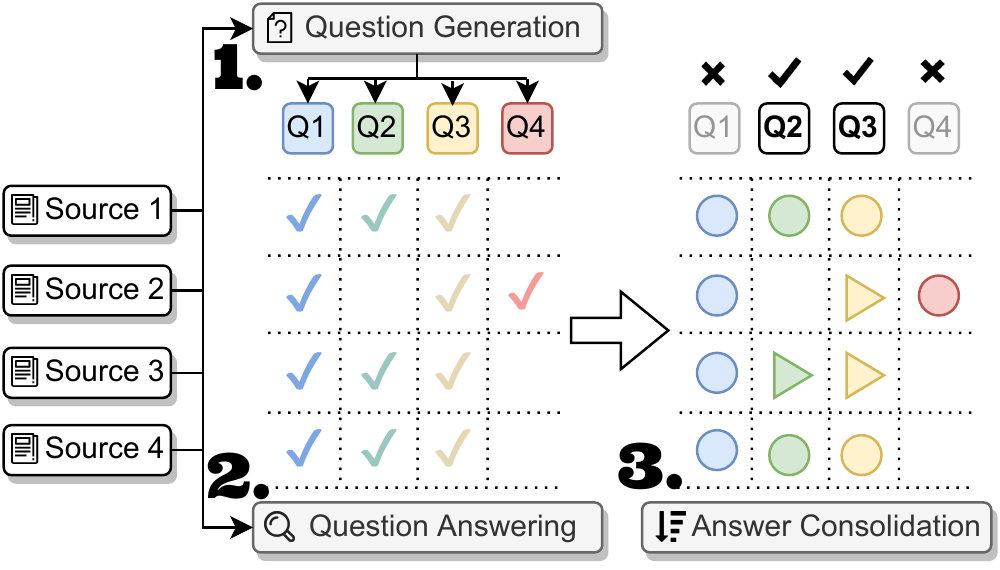}
    \caption{\textbf{Overview of the Discord Questions Framework.} The pipeline consists of: (1) question generation, (2) question answering, and (3) question consolidation, to find questions that news sources answer differently.}
    \label{fig:overall_framework}
\end{figure}

We hypothesize that a well-phrased question and a consolidated set of answers from the sources can reveal the coverage diversity of a story in a flexible and interpretable way for end-users. In our work, we operationalize the Discord Questions framework into a pipeline with three main components as shown in Figure~\ref{fig:overall_framework}. 
More specifically, we focus on two tasks: answer consolidation for the news domain and discord question generation. We create evaluation settings for each, allowing us to build high-performing models to use in a prototype implementation of the framework.

For answer consolidation, we repurpose existing QA evaluation works~\cite{chen2020mocha}, adapting it to achieve a balanced accuracy above 80\% on our built test set.
For discord question generation, we train a question generation model that improves the percentage of generated discord questions by 5\% compared to a strong baseline. We however estimate that our best-performing model still lags human-written question quality by at least 15\% in our evaluation setting.

We prototype the Discord Questions framework in a live demonstration. We rely on the Google News aggregator to obtain a listing of sources that cover a story and use our pipeline to generate several discord questions. Manual inspection reveals that questions generated by our system are found to be more interesting than other types of questions (such as factoid questions) and that the consolidated answers help surface diversity in terms of the level of detail, answer aspects, sentiment, and reasoning of sources, successfully revealing differences in coverage from news sources.

\section{Framework Definition}

We first define terminology, then introduce components of the discord questions framework.

\subsection{Terminology}

A news \textit{story} (sometimes \textit{topic} or \textit{event}) is a group of news articles published around the same time that discuss a common event and set of entities. Individual news \textit{articles} of a story are each published by a \textit{source}, a media organization that often hosts the article on its distribution platform. An article is composed of a headline, the article's \textit{content}, and optionally a summary. We denote the collection of articles' contents as the \textit{full context} of a story.

\subsection{Discord Question Pipeline}

The pipeline is visually summarized in Figure~\ref{fig:overall_framework}. It takes as input a story's news articles and follows three steps: (1) \textit{question generation} in which candidate discord questions are generated, (2) \textit{question answering} in which answers to a question are extracted from each source's content, and (3) \textit{answer consolidation}, in which a question's extracted answers are organized into semantic groups. The output is a set of questions and corresponding answer groups, which can be used to surface news coverage diversity.

\subsection{Discord Question Generation}

Discord Question Generation consists of using any of the sources' content to generate a question satisfying two properties: (1) \textbf{high coverage}, with most of the sources providing an answer to the question, (2) \textbf{answered diversely}, with answers exhibiting semantic diversity which can be organized into semantic groups.
We define cutoffs that assess if each property is respected. For property (1), the question should be answered by 30\% or more of the sources. For property (2), when grouping a question's answers, the largest group should contain no more than 70\% of all answers.

In Figure~\ref{fig:overall_framework}, out of the 4 candidate questions, only Q2 and Q3 satisfy both properties and are considered discord questions. Questions such as Q1 -- breaks property (2) -- are labeled as \textit{consensus questions}, as a majority of the sources' answers are in the same semantic group (i.e., circles). Factoid questions tend to be consensus questions (e.g., Who is the president of France?). Questions such as Q4 -- breaks property (1) -- are labeled as \textit{peripheral questions}, as a minority of sources answer the question. We hypothesize that consensus and peripheral questions are not pertinent to the study of a story's coverage diversity, as they do not reveal dimensions of source discord.
Section~\ref{section:dis_qg} explores ways to automatically generate discord questions.

\subsection{Question Answering}

Once a candidate question is generated, the question answering (QA) module extracts each source's answer -- if any -- to the question.

We leverage two properties of QA models in the Discord Questions framework. First, the QA model we use is extractive, selecting spans of text in the source's content that most directly answer the question without modification. Second, the model discerns when a source does not contain any answer to a question, predicting a \texttt{No Answer} special token.


In this work, we use a standard QA model, a RoBERTa-Large trained on common extractive QA datasets (details in Appendix~\ref{appendix:qa_model_details}), and reflect on the choice of QA model in the Limitations section.

\subsection{Answer Consolidation}

Once a question's answers are extracted, the final step is answer consolidation \cite{zhou2022answer}. The objective is to organize answers into semantic groups, with answers within the same group conveying semantically similar answers.

We follow \citet{zhou2022answer} and decompose answer consolidation into two sub-tasks: (1) answer-pair similarity prediction (also \textit{answer equivalence}), in which a model is tasked with assessing the similarity $S_{12}$ between two answers $(a_1, a_2)$ to a question $Q$, (2) the consolidation step, in which given a set of answers $(a_1, a_2, ..., a_n)$ and all pairwise similarities $S_{12}, S_{1n}, S_{2n}, ...$, the model must organize the answers into semantic groups.

Because answer-pair similarity can involve subjective opinion, \citet{chen2020mocha} framed the task as a regression problem, collecting human annotations on a 5-point Likert scale. \citet{bulian2022tomayto} later simplifies the task by framing it as binary classification and still achieve high inter-annotator agreement. We adopt the binary classification framing, as it simplifies annotation procedures.
In Section~\ref{section:answer_consolidation}, we collect an evaluation set for news answer consolidation and explore diverse transfer learning strategies, finding resources to build high-performing models for our application.

\section{Related Work}

\textbf{Analysis of media diversity} and bias often attempts to examine news coverage based on the media organizations that own the sources \cite{hendrickx2020dissecting}. The objective can be to map a source onto a left-right political range \cite{baly2018predicting}, or geopolitical origin (e.g., country) \cite{Hamborg2018BiasawareNA}. Information about source bias can be conveyed to the user through clustering \cite{park2009newscube} or matrix visualization \cite{Hamborg2018BiasawareNA}. Prior work has however shown that using visualization to increase news reader awareness can be challenging \cite{Spinde2020EnablingNC}. In the Discord Questions framework, we envision a new approach to news coverage diversity by revealing concrete examples of questions and organized answer groups that reveal source alignments.

\textbf{Answer Equivalence \& Consolidation.} Pre-trained models and large datasets have boosted QA performance, yet shallow metrics -- exact match and token F1 -- remain the most popular to assess performance \cite{chen2019evaluating}. Recent work on answer equivalence, MOCHA \cite{chen2020mocha} and Answer Equivalence (AE) \cite{bulian2022tomayto}, build methods to improve QA evaluation by manually collecting datasets of semantic similarities between reference and system answers to a question. \citet{zhou2022answer} formulate the task of answer consolidation and collect a large dataset to explore model performance on the task in the domain of online forums (i.e. Quora). In our work, we frame the answer consolidation task in the news domain and re-purpose answer equivalence models to achieve high performance on the task.

\textbf{Question generation} has expanded from the answer-aware sequence-to-sequence task \cite{du2017learning} to include many domain-specific applications, from  clarification QG \cite{rao2018learning}, inquisitive QG during a reading exercise \cite{ko2020inquisitive}, for conversation recommendations \cite{laban2020s}, factual consistency evaluation in summarization \cite{fabbri2021qafacteval} or to decompose fact-checking claims \cite{chen2022generating}. With discord questions, we add a new practical application of QG, to enable analysis of news coverage diversity.

\textbf{Multi-document summarization} (MDS), applied to product reviews \cite{di2014hybrid, bravzinskas2021learning} or in the news domain \cite{fabbri2019multi}, can be seen as related to discord questions. In MDS, models learn from the dataset content selection techniques, and whether to include or omit discordant information. Discord questions can be seen as targeted MDS focusing on story elements that involve source disagreement.

\section{News Answer Consolidation}
\label{section:answer_consolidation}

We collected an evaluation set we name NAnCo (\textbf{N}ews \textbf{An}swer \textbf{Co}nsolidation) and evaluated several transfer learning strategies to select the best-performing model for the pipeline.

\subsection{NAnCo Data}

To build a challenging evaluation set, we used a manual process to select questions and source answers for annotation.
At the time of annotation, we selected a hundred large stories in the recent section of Google News. Although Google News most likely applies a filter on the stories that appear in the recent section, we did not curate story selection beyond selecting stories with at least 25 sources. For each story, we use a baseline QG model -- a BART-large model \cite{lewis2020bart} trained on NewsQA \cite{trischler2017newsqa} -- to generate several thousand candidate questions. We then use a QA model to question answers from the story's full context. We filter to questions with at least 25 answers and manually select eight questions for which preliminary inspection reveals discord. In addition, we ensured that selected questions represented diverse topics (e.g., geopolitics, business, science), and structures (e.g., Why, How, What, and Who questions).

Statistics of NAnCo are summarized in Appendix~\ref{table:nanco_statistics}. For each question and answer set, we tasked three human annotators with grouping the answers semantically. The annotators were first shown an example question with pre-annotated groups by an author of the paper and could discuss the task before beginning annotation. Instructions given to the annotators are listed in Appendix~\ref{appendix:nanco_instructions}.

We follow \citet{laban2021news}'s procedure to aggregate multiple grouping annotations into \textit{global groups}, using a combination of majority voting and graph-based clustering \cite{blondel2008fast}. We then measure inter-annotator agreement using the Adjusted Rand Index measure between each annotator and a leave-one-out version of the global groups, and find an overall agreement of 0.76, confirming that consensus amongst annotators is high.

In the final dataset, questions have an average of 9.4 answer groups (ranging from 5-12), each with an average of 3.0 distinct answers per group (ranging from 1-25). We separate questions into two groups: four questions to a validation set available for hyper-parameter tuning, and four to a test set.

\subsection{Experimental Setting}

To facilitate experimentation, we convert final group labels into a binary classification task on pairs of answers. For each question, we look at all pairs of answers, assigning a label of $1$ if the two answers are in the same global group, and $0$ otherwise. In total, we obtain 3,267 pairs, with a class imbalance of 25\% of positive pairs.

The NAnCo data is large enough for evaluation, but too small for model training. We explore the re-use of existing resources to assess which transfers best to our task, specifically looking at models from NLI, sentence similarity, and answer equivalence.

For NLI models, we explore two models: \textbf{Rob-L-MNLI}, a RoBERTa-Large model \cite{liu2019roberta} trained on the popular MNLI dataset \cite{williams2018broad}, and \textbf{Rob-L-VitC} trained on the more recent Vitamin C dataset \cite{schuster2021get}, which has shown promise in other semantic comparison tasks such as factual inconsistency detection \cite{laban2022summac}. Model prediction is:
\begin{equation}
    S_{NLI}(A_1, A_2) = P(E \vert A_1, A_2) - P(C \vert A_1, A_2)
\end{equation}
Where $P(E \vert ...)$ and $P(C \vert ...)$ are model probabilities of the entailment and contradiction class. During validation, minor modifications such as a symmetric scoring, and using only $P(E \vert ...)$ had negligible influence on overall performance.

We explore two sentence embeddings models, selected on the Hugging Face model hub\footnote{\url{https://huggingface.co/models}} as strong performers on the Sentence Embedding Benchmark\footnote{\url{https://seb.sbert.net}}. First, \textbf{BERT-STS}, a BERT-base model \cite{devlin2018bert} finetuned on the Semantic Text Similarity Benchmark (STS-B) \cite{cerasemeval}. Second, \textbf{MPNet-all}, an MPNet-base model \cite{song2020mpnet} trained on a large corpus of sentence similarity tasks \cite{reimers-2019-sentence-bert}.

Finally, we select four answer equivalence models. First, \textbf{LERC} is a BERT-base model introduced in \citet{chen2020mocha}. Second, \textbf{Rob-L-MOCHA}, a RoBERTa-Large model trained on MOCHA's regression task, which requires predicting an answer pairs similarity on a scale from 1 to 5. Third, \textbf{Rob-L-AE}, a RoBERTa-Large model we train on the AE's binary classification task which determines whether an answer pair is similar or not. Fourth, the \textbf{RobL-MOCHA-AE} model, which we train on a union of MOCHA and AE, adapting the classification labels to regression values (i.e., label 1 to value 5, label 0 to value 0).

We note that not all models have access to the same input. NLI and Sentence Embeddings models are not trained on tasks that involve questions, and we only provide answer pairs for those models. Answer equivalence-based models see the question as well as the answer pair, as prior work has shown that it can improve performance \cite{chen2020mocha}.

All models produce continuous values as predictions. The threshold for classification is selected on the validation set, and used on the test set to assess realistic performance. Technical details for training and usage of the eight models are in Appendix~\ref{appendix:nanco_models}.

\subsection{Results}

\begin{table}[]
    \centering
    \resizebox{0.44\textwidth}{!}{%
    \begin{tabular}{lcccc}
    & \textbf{MOCHA} & \textbf{AE} & \multicolumn{2}{c}{\textbf{NAnCo}} \\
    \cmidrule(lr){2-2} \cmidrule(lr){3-3} \cmidrule(l){4-5}
    \textbf{Model Name} & Test & Test & \textbf{Valid.} & \textbf{Test} \\
    \cmidrule(r){1-1} \cmidrule(lr){2-2} \cmidrule(lr){3-3} \cmidrule(l){4-5}
    Rob-L-MNLI & 0.07  & 54.7 & 67.6 & 58.1 \\
    Rob-L-VitC & -0.01 & 54.7 & 69.7 & 69.7 \\
    \cmidrule(r){1-1} \cmidrule(lr){2-2} \cmidrule(lr){3-3} \cmidrule(l){4-5}
    BERT-STS  & 0.70 & 80.0 & 84.7 & 73.3 \\
    MPNet-all & 0.61 & 75.7 & 85.4 & 73.0 \\
    \cmidrule(r){1-1} \cmidrule(lr){2-2} \cmidrule(lr){3-3} \cmidrule(l){4-5}
    LERC        & 0.81 & 82.2 & 87.5 & 70.9 \\
    Rob-L-MOCHA & \textbf{0.87} & 84.5 & \textbf{92.9} & \textbf{81.3} \\
    Rob-L-AE    & 0.61 & \textbf{89.9} & 73.5 & 64.6 \\
    Rob-L-MOCHA-AE & \textbf{0.87} & 89.2 & 89.9 & 74.1 \\
    \bottomrule
    \end{tabular}
    }
    \caption{\textbf{Results on MOCHA (correlation), Answer Equivalence (balanced acc.), and NAnCo (balanced acc.).} Eight models were tested: NLI (top 2), sentence embeddings (middle 2), answer equivalence (bottom 4).}
    \label{table:nanco_results}
\end{table}

In Table~\ref{table:nanco_results}, we report Pearson correlation scores for MOCHA, and balanced accuracy for AE and NAnCo to account for class imbalance.

On all datasets, answer equivalence models perform best, followed by sentence embeddings models, and NLI models perform worst.  Within answer equivalence models, Rob-L-MOCHA tops performance, outperforming both LERC -- a smaller model trained on the same data -- and AE-trained models. We hypothesize that the more precise granularity of MOCHA provides additional signals useful to our task. Surprisingly, training on the union of MOCHA and AE does not improve performance, hinting at differences between the datasets, and a closer resemblance of our task to MOCHA.

All models see a decrease in performance when transitioning from validation to test settings. This drop in performance reflects the reality of using models in practice, in which a threshold must be selected in advance.

Although a test balanced accuracy of 81.3\% is far from errorless, the performance is encouraging and we use Rob-L-MOCHA when assembling the framework in Section~\ref{section:assembled_framework}. In practice, for a set of answers to a question, we run Rob-L-MOCHA on all answer pairs, build a graph based on predictions, and run the Louvain clustering algorithm \cite{blondel2008fast} to obtain answer groups.

\section{Discord Question Generation}
\label{section:dis_qg}

\begin{figure}
    \centering
    \includegraphics[width=0.44\textwidth]{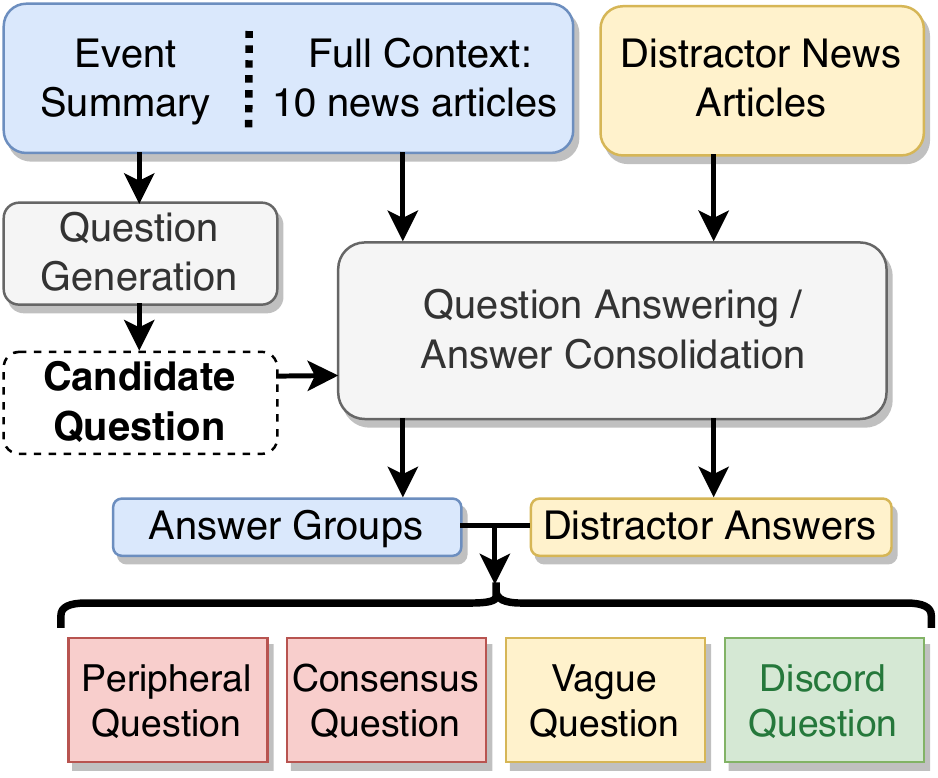}
    \caption{\textbf{Diagram of automatic evaluation of question candidates.} Questions are tagged into one of four categories: peripheral, factoid, vague and discord.}
    \label{fig:disqg_diagram}
\end{figure}

The Discord Question framework relies on obtaining story-relevant questions. QG models are known to excel at generating factoid questions but are limited on realistic curiosity-driven questions \cite{scialom2020ask}.
We propose an automatic method to evaluate QG models on the ability to generate discord questions, based on the intuition that we can use a story's full context to evaluate a question. The method is illustrated in Figure~\ref{fig:disqg_diagram}.

\subsection{Evaluation Method}

We select 200 news stories on the recent section of Google News, omitting stories with less than 10 sources. For each, we extract the \textit{full context}, as well as a summary selected from one of the articles.

All QG models receive the summary and generate a candidate question. Crucially, models do not have access to the full context but must generate questions with diverse answers in the full context.

Once a candidate is generated, the QA module extracts all potential answers ($A$) to the question from the full context, and the answer consolidation module groups answers semantically.
If no answer is extracted, or answers were extracted from fewer than 30\% of sources, we label the question as a \textbf{peripheral} question.
If answer consolidation finds that a single answer group accounts for at least 70\% of answers, we label the question as a \textbf{consensus} question. We find that factoid questions often fall in this category (e.g., \textit{Who is the president of X?}).

We note that the thresholds set to filter out peripheral and consensus questions were chosen empirically are can be modified depending on the application setting. For example, regarding the threshold for labeling a question as peripheral, lowering the threshold leads to producing more discord questions, including more specific questions that are not central to the story, while increasing the threshold would lead the pipeline to produce no discord questions.

A common limitation of QG is a preference for vague and common questions \cite{heilman2010good}, a problem that exists in other NLG domains such as dialogue response generation \cite{li2016diversity}. With overly vague questions (e.g., \textit{What did they say?}), models increase the likelihood of being answered. Vague questions are undesirable in our framework, as differing answer groups might arise not from discord, but from ambiguity.

We devise an automatic method to detect vague questions, borrowing from the concepts of TF-IDF and term specificity \cite{jones1972statistical}. We use 10 \textit{distractor} news articles published several months before the news story. For a candidate question, we extract all answers to the question from the distractor articles $(A_{dis})$. We compute a question specificity score as:
\begin{equation}
    Spec(Q, A, A_{dis}) = \frac{\vert A \vert}{\vert A_{dis} \vert +\epsilon},
\end{equation}
where we set $\epsilon=0.001$ for numerical stability, and $\vert A\vert$ is the number of answers. If there are few distractor answers, specificity is large, otherwise, if $Spec(Q, A, A_{dis}) \leq 2$, we label the question as \textbf{vague}. Other candidates are labeled as \textbf{discord} questions, as they (1) are answered by a large proportion of sources, (2) have several groups of answers, and (3) are specific to the story.

\begin{table}[]
    \centering
    \resizebox{0.47\textwidth}{!}{%
    \begin{tabular}{lccccc}
     & \multicolumn{5}{c}{\textbf{\% Discord Questions}} \\
    \cmidrule(l){2-6}
    \textbf{Model-Dataset} & \textbf{How} & \textbf{Why} & \textbf{What} & \textbf{Who} & \textbf{Avg.} \\
    \cmidrule(l){1-1} \cmidrule(l){2-5} \cmidrule(l){6-6}
    BART-SQuAD & 19 & 63 & 13 & 5 & 25.0 \\
    T5-SQuAD & 12 & 63 & 18 & 8 & 25.3 \\
    MixQG-SQuAD & 11 & 62 & 27 & 9 & 27.3 \\
    \cmidrule(l){1-1} \cmidrule(l){2-5} \cmidrule(l){6-6}    
    BART-NewsQA & 3 & \textbf{68} & 38 & 7 & 29.0 \\
    T5-NewsQA & 2 & 65 & 42 & 8 & 29.3 \\
    MixQG-NewsQA & 6 & 66 & 42 & 8 & 30.5 \\
    \cmidrule(l){1-1} \cmidrule(l){2-5} \cmidrule(l){6-6}
    BART-Fairy & 31 & 54 & 60 & 3 & 37.0 \\
    T5-Fairy & 42 & 63 & 58 & 6 & 42.3 \\
    MixQG-Fairy & 33 & 61 & 49 & 11 & 38.5 \\
    \cmidrule(l){1-1} \cmidrule(l){2-5} \cmidrule(l){6-6}
    BART-Inqui & 43 & 65 & 42 & 13 & 40.8 \\
    T5-Inqui & 46 & 58 & 43 & \textbf{14} & 40.3 \\
    MixQG-Inqui & 37 & 50 & 34 & 13 & 33.5 \\
    \cmidrule(l){1-1} \cmidrule(l){2-5} \cmidrule(l){6-6}
    T5-Discord & \textbf{49} & 64 & \textbf{65} & \textbf{14} & \textbf{48.0} \\
    \cmidrule(l){1-1} \cmidrule(l){2-5} \cmidrule(l){6-6}
    Human Written & \textbf{73} & \textbf{87} & \textbf{66} & \textbf{27} & \textbf{63.0} \\
    \bottomrule{}
    \end{tabular}
    }
    \caption{\textbf{Results on Discord QG.} For each model (BART, T5, MixQG), and dataset (SQuAD, NewsQA, Fairy, and Inqui) we report the \% of questions tagged as discord. T5-Discord is the model trained on data we curate, and we report a human performance estimate.}
    \label{table:disqg_results}
\end{table}

\subsection{Experimental Setting}
\label{section:qgen_experiment}

\begin{figure*}
    \centering
    \includegraphics[width=0.98\textwidth]{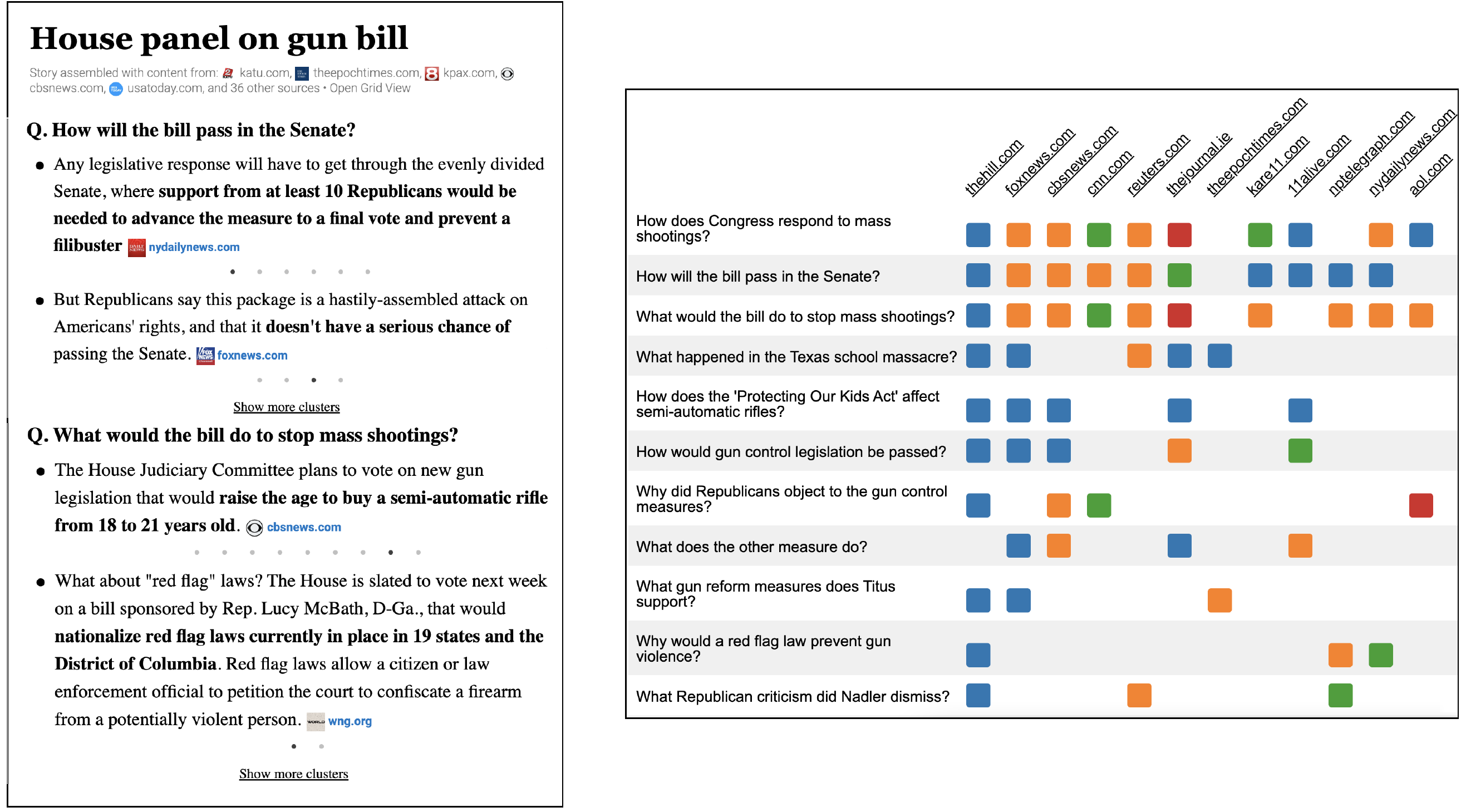}
    \caption{\textbf{Prototype interface of the Discord Questions demonstration.} The Q\&A view (left) lists the most covered discord questions and answers. The Grid view (right) condenses the story information in a matrix.}
    \label{fig:disq_interface}
\end{figure*}

We experiment with three models: BART-large, T5-large \cite{raffel2020exploring}, and MixQG-large \cite{murakhovska2022mixqg}, a model designed for QG. For each, we finetune on four datasets: SQuAD, NewsQA, FairyTaleQA \cite{xu2022fantastic} which has narrative comprehension questions, and InquisitiveQG \cite{ko2020inquisitive} which collected questions readers think of while reading.

A confounding factor in QG is the choice of start word. Start words affect the difficulty of generating discord questions, with a difference between words that more often lead to factoid questions (e.g., Where), or reasoning starting words (e.g., Why). A model that generates a larger fraction of Why questions might be advantaged, regardless of its ability on all start words. To counter the start word's effect, we enforce that models are compared using the same start words.

For each of the 200 test stories, models generate one question for four start words: Why, How, What, and Who (we skip Where and When as our observation revealed a very low percentage of discord questions), for a total of 800 candidate questions.

To understand task feasibility, we collect human-generated discord questions. We manually wrote a candidate discord question for each story and start word combinations. Although not necessarily an upper bound of performance, it can serve as a rough estimate of human performance.

\subsection{Results}

Results for QG models and human performance in Table~\ref{table:disqg_results}. Overall, human performance outperforms models by a large margin for all start words. As expected, the start word affects task difficulty, with discord percentages lower for Who questions, even in the human-written condition.

The dataset influences performance more than model choice, and in particular different datasets lead to the best performance on different start words. For example, NewsQA models achieve the highest performance on the Why questions, FairyTale models on the What questions, and Inquisitive models on the How and Who questions. This insight leads us to aggregate a Discord dataset by concatenating: \texttt{(Inqui/How, NewsQA/Why, FairyTale/What,} and \texttt{Inqui/Who)}. We train a T5-large model on Discord, and achieve the highest overall performance of 48\% discord questions generated, an absolute improvement of 5.7\%, even though performance still lags human-written questions by around 15\%.

Automatic evaluation is inherently limited. We next complement our results with manual annotation of generated discord questions.

\section{Discord Questions Assembled}
\label{section:assembled_framework}

We assemble the Discord Questions frame with the best-performing components -- \texttt{Rob-L-MOCHA} for consolidation and \texttt{T5-Discord} for QG -- and make a public web interface. We perform a manual evaluation of the system, first evaluating the relative interestingness of discord questions to potential readers, and second analyzing types of diversity surfaced by the system.

\subsection{Assembly Detail}
\label{section:assembly_detail}

In the demonstration, we collect stories as they are added to the English version of Google News, filtering to stories with at least 10 distinct sources. For each story, we obtain article content using the \texttt{newspaper} library \footnote{\href{https://github.com/codelucas/newspaper}{github.com/codelucas/newspaper}} and run the Discord Questions pipeline, often generating several hundred candidate questions and filtering down to discord questions that receive highest source coverage.

We design two interface to visualize stories: \textit{Q\&A view} and \textit{Grid view}, both shown in Figure~\ref{fig:disq_interface}. In the Q\&A view, the user sees a list of selected discord questions and a horizontal carousel with a representative answer from each answer group. Sources are linked explicitly, the user can click to access the original article. In the Grid view, information is condensed into a matrix to facilitate comparison between sources: each row lists a question, each column represents a source, in each entry, a shape indicates whether a source answered a question, and the shape's color indicates the source's answer group.

Work on the interface is preliminary, and serves as a proof of concept, demonstrating it can be run on several hundred stories a day with a moderate amount of infrastructure. Further investigation through usability studies is required to understand the usefulness of the framework to news readers.

\subsection{Are Discord Questions Interesting?}
\label{section:preference_study}

The automatic evaluation in Section~\ref{section:qgen_experiment} does not consider the interestingness of the question: a question might qualify as a discord question while not covering an interesting aspect of the story for the reader. Interest in a question is inherently subjective, and we perform a manual annotation of questions to evaluate the relative interestingness of discord questions to other question categories.

We randomly select 300 question pairs from Section~\ref{section:qgen_experiment}'s experiment. Each pair contains one question marked as discord, and one marked as any other category (i.e., peripheral, consensus, vague) for the same story. Three annotators read the shuffled question pair ($Q_1$, $Q_2$) and optionally the story's summary, and select the question they would be more interested in seeing answered.
The annotator can choose: $Q_1$ wins, $Q_2$ wins, both are not interesting, or both are interesting. Appendix~\ref{appendix:q_pref_instructions} relays task instructions and detailed results.

We compute the inter-annotator agreement level through Cohen's Kappa, and find an agreement level of 0.51 or moderate agreement, confirming that though interest in a question is subjective, some agreement amongst annotators exists. We find that annotators preferred discord questions in 68\% of cases, confirming that discord questions are not only relevant for surfacing diversity in source coverage, but they also are more interesting to news readers. We note that a preference in 68\% of cases shows that in many cases, consensus and peripheral questions are interesting as well, and discord questions are one of many ways to generate interesting questions in news reading applications.

\subsection{Types of Diversity Surfaced}

To gain an understanding of the types of surfaced diversity, we inspect discord questions generated by our pipeline from 32 Google News stories from the Business, World, and Science sections. For each story, we annotate up to five questions with at least 3 answer groups, annotating 100 questions.

We annotated each question with whether the question qualifies as a discord question, and if so, the type of diversity it reveals. We found that 16\% of questions are erroneously tagged as discord and the remaining 84\% surface four types of diversities.

Causes for errors were: (1) the question is vague and sources answer different question interpretations (14\%), and (2) answers are all semantically similar but the consolidation module mistakenly creates multiple groups (2\%).

For valid discord questions, we labeled each with up to four types of coverage diversity it reveals, expanding on prior work in answer equivalence \cite{bulian2022tomayto}:
\begin{itemize}
    \setlength\itemsep{-0.5em}
    \item \textbf{Level-of-detail Difference.} 79\% of valid questions surface differences between coarse and precise answers to the question,
    \item \textbf{Aspect Difference.} 66\% bring to light differences in the aspects answers focus on (e.g. economics vs. politics),
    \item \textbf{Sentiment Difference.} 41\% reveal source answers being more positive, neutral, or negative towards a question,
    \item \textbf{Reason Difference.} 22\% expose differences in the reason or prediction a source makes about a question.
\end{itemize}

See Appendix~\ref{appendix:example_diversity} for examples of each type of diversity. From this analysis, we conclude that although the pipeline produces errors, the majority of generated questions reveal some coverage diversity.

\section{Conclusion}

We introduce the Discord Questions framework, in which we hypothesize that a question accompanied by an organized set of answers from the sources can spotlight specific ways in which sources disagree, providing concrete examples of coverage diversity to a news reader.
We decompose the framework into required components, and design evaluation methodology for each. We select high-performing models for each component and assemble them into a working prototype of the framework. We confirm through manual analysis that questions generated within our framework are of interest to potential users, and in a majority of cases surface four types of diversity in news coverage, from varying levels of detail in the reporting, to differing sentiment or reasoning about an event's cause, confirming that discord questions are an interpretable tool to uncover coverage diversity.


\section{Limitations}
\label{section:limitations}

In this work, we focus on generating discord questions, filtering out other types of questions. Discarded questions can however be valued in other settings, and our selection process should not be seen as a general assessment of question quality. For instance, \citet{fabbri2021qafacteval} show that generating highly specific factoid questions can boost performance in factual inconsistency detection in summarization, and unanswered questions help challenge QA models \cite{rajpurkar2018know}.

Our demonstration relies on components susceptible to making errors, which can compound as one module's errors are forwarded to the next. For example, an inaccurately extracted answer by the QA model will lower the quality of answer groups in the consolidation step. In particular, extractive QA models can be limiting when answers are indirect or implied \cite{chen2022generating}. On the bright side, modularity enables us to swap to improved components, for instance as generative QA becomes available \cite{tafjord2021general}.

The framework we propose assumes that exposing a news reader to coverage from a diverse set of sources is beneficial, however, exposure to media bias can be detrimental, in some cases misrepresenting important geopolitical events such as elections \cite{allcott2017social} and wars \cite{kuypers2006bush}. Therefore, a careful balance is required in source selection to present diverse perspectives to the user, while not promoting dangerous misrepresentations. In the implementation presented in Section~\ref{section:assembled_framework}, we rely on Google News' source selection process \footnote{\href{https://developers.google.com/search/blog/2021/06/google-news-sources}{developers.google/blog/2021/06/google-news-sources}}, which accounts for transparency and editorial practices of a source. Google News is however not a gold standard, as it is known to have Western bias \cite{Watanabe2013TheWP} and the aggregator recently removed major Russian sources from its platform\footnote{\href{https://www.reuters.com/technology/exclusive-google-drops-rt-other-russian-state-media-its-news-features-2022-03-01/}{reuters.com/technology/google-drops-rt-other-russian}}.

Our current prototype is inherently limited due to our focus on English-written news, as coverage diversity on international topics is likely to come from non-English news sources. However, improvements in automatic news translation \cite{tran2021facebook}, as well as multi-lingual models \cite{hu2020xtreme} draw a path towards a multi-lingual version of our prototype.

As stated earlier, the prototype interface we built remains work-in-progress, and usability studies -- planned as future work -- are required to examine the effect of discord questions on readers' understanding of the news. Furthermore, beyond the setting of Google News stories, discord questions could be beneficial on the study of long, unfolding news stories \cite{laban2017newslens}, helping readers form evolving opinions over time. Finally, future work should aim to integrate discord questions into non-standard news interfaces such as chatbots \cite{laban2020s} or podcasts \cite{laban2022newspod}.

\section{Ethical Consideration}

We focused our experiments for the Discord Questions on the English language, and even though we expect the framework to be adaptable to other languages and settings, we have not verified this assumption experimentally and limit our claims to the English language.

The models and datasets utilized primarily reflect the culture of the English-speaking populace. Gender, age, race, and other socio-economic biases may exist in the dataset, and models trained on these datasets may propagate these biases. Question generation and answering tasks have previously been shown to contain these biases.

We note that the models we use are imperfect and can make errors. When interpreting our framework's outputs, results should be interpreted not in terms of certainty but probability. For example, if our system states that a source did not answer a specific discord question, there is a probability that the source answered the question, but the question-answering module we use failed to extract such an answer.

To build the components of our prototype, we relied on several datasets as well as pre-trained language models. We explicitly verified that all datasets and models are publicly released for research purposes and that we have proper permission to reuse and modify the models.

\bibliography{anthology,custom}
\bibliographystyle{acl_natbib}

\newpage
\appendix

\section*{Appendix}
\renewcommand{\thetable}{A\arabic{table}}
\setcounter{table}{0}
\renewcommand{\thefigure}{A\arabic{figure}}
\setcounter{figure}{0}

\section{QA Model Details}
\label{appendix:qa_model_details}

We use a RoBERTa-Large model as the basis of the extractive QA component of the Discord Questions pipeline. We finetune the model on a combination of two common QA datasets: SQuAD 2.0 \cite{rajpurkar2018know}, and NewsQA \cite{trischler2017newsqa}. We use the Adam optimizer for training with a learning rate of $2*10^{-5}$ and a batch size of 32. Hyper-parameters were selected through tuning on the validation set of the datasets. The final model checkpoint achieves an F1 score of 86.7 on the SQuAD 2.0 test set, and 68.9 on NewsQA's test set, within a few points of previous results \cite{liu2019roberta}.

\section{NAnCo Annotation Instructions}
\label{appendix:nanco_instructions}

The instructions that were given to the three annotators are listed in Figure~\ref{fig:nanco_instructions}. As listed in the instructions, the annotators were tasked with first looking through an annotated example before starting the annotation. One annotator asked clarification questions about the use of ``-1'' annotations before proceeding.

\begin{figure}
    \begin{framed}{Welcome!
    
        This sheet contains 9 tabs (Q1-Q9), each containing a question (at the top) and 25-50 answers to that question from different sources.
        The goal is to annotate, in each tab, which answers give the same answer elements (are in the same cluster).
        
        In each tab, you should fill out the Cluster Annotation column.\\
          - For each answer row, the cluster annotation should be a number, such that all the answers that you believe give the same answer should receive the same cluster number.\\
          - The cluster numbers do not need to be consecutive (for instance if you change your mind about a cluster)\\
          - You can move the answer rows around if you want to (for example put similar answer rows next to each other), but it is not necessary. \\
          - If you believe an answer does not contain a valid answer, you should annotate that with a "-1". These will be removed from annotation and not considered an answer cluster. \\
        
        The first tab Q1 has already received annotation. Review the sheet's annotation, and if you disagree, or want to discuss an annotation choice, reach out to Philippe to discuss. If you have other questions about the task, reach out as well.
        
        Once you feel like you understand the task, feel free to start annotation. We anticipate the task to take 2-3 hours to annotate the 8 spreadsheets (Q2-Q9).}
    \end{framed}
    \caption{Instructions for the annotation of the NAnCo evaluation set.}
    \label{fig:nanco_instructions}
\end{figure}

\section{NAnCo Statistics}
\label{appendix:nanco_statistics}
Table~\ref{table:nanco_statistics} lists the eight questions included in the NAnCo evaluation we created, with the first four questions in the validation set, and the last four in the test set.

\begin{table}[]
    \centering
    \resizebox{0.47\textwidth}{!}{%
    \begin{tabular}{p{4.5cm}cccc}
    \toprule 
    \textbf{Question} & \textbf{\#Ans} & \textbf{\#Clus} & \textbf{\#Pairs} & \textbf{IAA} \\
    \hline 
    Why did Governor Abbott order additional inspections? & 29 & 8 & 406 & 0.95 \\
    How long will cocktails to-go be around? & 28 & 10 & 378 & 1.0 \\
    How would Australia support the Solomon Islands? & 26 & 12 & 325 & 0.57 \\
    What kind of relationship does Musk have with Twitter? & 31 & 11 & 465 & 0.68 \\
    \hline 
    \hline 
    What caused Delta shares to rise? & 26 & 7 & 351 & 0.87 \\
    What do astronomers consider the Oort Cloud to be? & 26 & 11 & 325 & 0.70 \\
    How does Biden handle inflation? & 24 & 11 & 276 & 0.50 \\
    Who would object to Sweden joining NATO? & 39 & 5 & 741 & 0.79 \\
    \hline 
    \textbf{Total} & \textbf{229} & \textbf{75} & \textbf{3267} & \textbf{0.76} \\
    \bottomrule 
    \end{tabular}
    }
    \caption{\textbf{Statistics of the NAnCo dataset.} Eight questions (top 4 in validation, bottom 4 in test set) are annotated. We report the number of answers, annotated clusters, samples in the pairwise classification task, and the annotator agreement level.}
    \label{table:nanco_statistics}
\end{table}

\section{NAnCo Model Details}
\label{appendix:nanco_models}

Reproducibility details of the eight models included in the NAnCo experiments:
\begin{enumerate}
    \item \textbf{Rob-L-MNLI}: Corresponds to \texttt{roberta-large-mnli} on the Hugging Face model hub \footnote{\url{https://huggingface.co/models}}
    \item \textbf{Rob-L-VitC}: Corresponds to \texttt{tals / albert-xlarge-vitaminc-mnli} on the Hugging Face model hub
    \item \textbf{BERT-STS}: Corresponds to the \texttt{sentence-transformers / stsb-bert-base} on the HuggingFace model hub
    \item \textbf{MPNet-all}: Corresponds to the \texttt{sentence-transformers / all-mpnet-base-v2} on the HuggingFace model hub
    \item \textbf{LERC}: Corresponds to the pre-trained model released by \citet{chen2020mocha}\footnote{\url{https://github.com/anthonywchen/MOCHA}}
    \item \textbf{Rob-L-MOCHA}: We train this model initializing with a RoBERTa-Large model, and training on the MOCHA dataset using an L2 regression loss, the Adam optimizer, a learning rate of $1*10^{-5}$, and a batch size of 10. The final checkpoint achieves a Mean Average Error (MAE) of 0.4322 on the validation dataset.
    \item \textbf{Rob-L-AE}: We train this model initializing with a RoBERTa-Large model, and training on the AE dataset using a cross-entropy loss, the Adam optimizing, a learning rate of $1*10^{-5}$, and a batch size of 32. The final checkpoint achieves an F1 of 90.9 on the validation set of AE.
    \item \textbf{Rob-L-MOCHA-AE}: We train this model initializing with a RoBERTa-Large model, and training on the combination of MOCHA and a regression version of AE, using an L2 regression loss, the Adam optimizer, a learning rate of $1*10^{-5}$, and a batch size of 10. The final checkpoint achieves a Mean Average Error (MAE) of 0.5288 on the validation dataset.
\end{enumerate}

\section{Preference Annotation Details}
\label{appendix:q_pref_instructions}

Figure~\ref{fig:q_pref_instructions} details the instructions that were given to the three annotators that participated in the preference selection task described in Section~\ref{section:preference_study}.

\begin{figure}
    \begin{framed}
    Welcome!\\
    On each row, read through the two questions. If it is unclear what news story it is about, you can read through the summary as an additional context (Note that it is ok if you can't find an answer from the summary given a question).\\
    The task consists of choosing which question you believe is more interesting and central to the story. That is, please select a question that you are more curious/willing to know what are the answers from different source articles.

    Options for preference are:\\
    - 1 (if you believe question1 is more interesting)\\
    - 2 (if you believe question2 is more interesting)\\
    - 0 (if both questions are roughly equally not interesting)\\
    - 3 (if both questions are roughly equally interesting)\\

    The first example (row 5) is labeled as an example (news story about Wikipedia and Bitcoin). The preference is set as 1 as the first question (How did Wikipedia's decision affect the free web-based encyclopedia?) is judged to be more interesting than question 2 (How long will Wikipedia stop accepting cryptocurrency donations? which asks about a detail that might not be stated).
    \end{framed}
    \caption{Instructions for the annotation of the preference over question interestingness}
    \label{fig:q_pref_instructions}
\end{figure}


\section{Coverage Diversity Category Examples}
\label{appendix:example_diversity}

\begin{table*}[]
    \resizebox{\textwidth}{!}{%

    \begin{tabular}{p{6cm} p{6cm} p{6cm}}
    \toprule \\
    \multicolumn{3}{c}{\Large{Vague Questions (14\%)}} \\
    \multicolumn{3}{c}{Where did the investigators get the data from?}\\\\

    Data reported \textbf{in Morbidity and Mortality Weekly Report} did not demonstrate an increase in pediatric hepatitis [...] [\url{healio.com}] & the link between hepatitis cases in children and COVID is inconclusive, but data \textbf{from all around the world} suggests it exists [\url{ynetnews.com}] & and about 14 percent of hospitalized patients are admitted into intensive care, the commission said, citing \textbf{research data from Europe}. [\url{ecns.cn}] \\
    \multicolumn{3}{c}{\textit{Reasoning: The sources are not discussing the same data as the question is vague.}} \\

    \\ \toprule \\

    \multicolumn{3}{c}{\Large{No Discord (2\%)}} \\
    \multicolumn{3}{c}{When did China have access to TikTok's database?}\\
    
    [...] between September 2021 and January 2022. [\url{businessinsider.com}] & & [...] between last September and January this year. [\url{asiafinancial.com}] \\
    \multicolumn{3}{c}{\textit{Reasoning: The relative and absolute dates refer to the same period, this is an error induced by the consolidation module.}} \\
    
    \\ \toprule \toprule \\

    \multicolumn{3}{c}{\Large{Level of Detail (79\%)}} \\
    \multicolumn{3}{c}{What was the TikTok report about?}\\

    the access that China still has over the US region [...] [\url{techtimes.com}] & China-based engineers working for Tiktok have repeatedly accessed TikTok's US user data [\url{theinformation.com}] & [...] audio from more than 80 internal meetings of the popular social media platform has been leaked, exposing Chinese-based TikTok employees repeatedly accessing US user data [\url{euroweeklynews.com}] \\
    
    \multicolumn{3}{c}{\textit{Reasoning: From left to right, the sources reveal a little, moderate, and fine level of detail in answering the question.}} \\
    
    \\ \toprule \\

    \multicolumn{3}{c}{\Large{Different Aspects (66\%)}} \\
    \multicolumn{3}{c}{Why would LEGO receive \$56 million?}\\

    The company will be eligible for [...] performance grants as part of the CommonWealth's Major Employment and Investment Program. [\url{washingtonian.com}] &  & Lego will be eligible to receive a grant [...] based on an investment of \$1 billion and the creation of jobs in excess of 1,760 [...] [\url{virginiabusiness.com}] \\

    \multicolumn{3}{c}{\textit{Reasoning: The left source answers from a political perspective, whereas the right source gives an economics perspective.}} \\

    \\ \toprule \\

    \multicolumn{3}{c}{\Large{Different Sentiment (41\%)}} \\
    \multicolumn{3}{c}{What impact does the strike action have on Network Rail?}\\

    as the national strike [...] is likely to cause disruption to travel.  [\url{stokesentinel.co.uk}] & The high-profile walk-outs will also have a \textbf{shattering knock-on impact} on local and national rail services [\url{grimsbytelegraph.co.uk}] & Rail services \textbf{have been ravaged} - down more than 80\% across the North West [\url{news.sky.com}] \\

    \multicolumn{3}{c}{\textit{Reasoning: Source on the left is neutral, and the middle and right sources gradually get stronger in sentiment.}} \\

    \\ \toprule \\    	

    \multicolumn{3}{c}{\Large{Different Reasons (22\%)}} \\
    \multicolumn{3}{c}{How did the Israeli observation balloon crash?}\\

    [...] an Israeli balloon crashed and fell in Northern Gaza Strip [\url{mymcmurray.com}] & [...]  the military said it became disconnected from its anchor for unknown reasons. [\url{timesofisrael.com}] & The Palestinian resistance in Gaza announced on Friday that it had shot down an Israeli military surveillance balloon [...] [\url{middleeastmonitor.com}] \\

    \multicolumn{3}{c}{\textit{Reasoning: Source on the left is not specific, but the middle and right sources give contradicting reasons for the balloon crash.}} \\

    \\ \bottomrule
    \end{tabular}
    }
    \caption{Examples of the two types of errors found in our manual analysis (Vague and No Discord Questions), as well as the four types of diversities surfaced by the discord questions.}
    \label{appendix:discord_typology_examples}
\end{table*}

\end{document}